\documentclass[11pt]{article}

\usepackage{graphicx, verbatim}
\usepackage{amssymb}
\usepackage{alltt}
\usepackage{amsmath, amssymb, amsfonts, amscd, xspace, pifont, amsthm}
\usepackage{mathrsfs}
\usepackage{algorithm}

\def\twoImages#1#2#3#4#5#6 
{
  \centerline{\hfill\makebox[#2]{#3}\hfill\makebox[#5]{#6}\hfill}
  \centerline{\hfill
    \includegraphics[width=#2]{#1}
    \hfill
    \includegraphics[width=#5]{#4}
    \hfill}
}

\newcommand{\F}[1]{\ensuremath{\mathrm{#1}}\xspace}

\usepackage{authblk}
\def\threeImages#1#2#3#4#5#6#7#8#9 
{
  \centerline{\hfill\makebox[#2]{#3}\hfill\makebox[#5]{#6}\hfill\makebox[#8]{#9}\hfill}
  \centerline{\hfill
    \includegraphics[width=#2]{#1}
    \hfill
    \includegraphics[width=#5]{#4}
    \hfill
    \includegraphics[width=#8]{#7}
    \hfill}
}

\DeclareMathOperator*{\argmin}{arg\,min}

\date{}
\title{Link prediction for partially observed networks}
\author[1]{Yunpeng Zhao}
\author[2]{Elizaveta Levina}
\author[2]{Ji Zhu}
\affil[1]{Department of Statistics, George Mason University, Fairfax, VA 22030 }
\affil[2]{Department of Statistics, University of Michigan,
Ann Arbor, MI 48109}

\begin{document}
\maketitle
\begin{abstract}
Link prediction is one of the fundamental problems in network analysis. In many applications, notably in genetics, a partially observed network may not contain any negative examples of absent edges, which creates a difficulty for many existing supervised learning approaches.  We develop a new method which treats the observed network as a sample of the true network with different sampling rates for positive and negative examples. We obtain a relative ranking of potential links by their probabilities, utilizing information on node covariates as well as on network topology.  Empirically, the method performs well under many settings, including when the observed network is sparse. We apply the method to a protein-protein interaction network and a school friendship network.    
\end{abstract}

\section{Introduction}
A variety of data in many different fields can be described by networks. Examples include friendship and social networks, food webs, protein-protein interaction and gene regulatory networks, the World Wide Web, and many others.

One of the fundamental problems in network science is link prediction, where the goal is to predict the existence of a link between two nodes based on observed links between other nodes as well as additional information about the nodes (node covariates) when available (see \cite{Lu&Zhou2010}, \cite{Liben2007} and \cite{Getoor2005} for recent reviews). Link prediction has wide applications. For example, recommendation of new friends or connections for members is an important service in online social networks such as Facebook. In biological networks, such as protein-protein interaction and gene regulatory networks, it is usually time-consuming and expensive to test existence of links by comprehensive experiments;  link prediction in these biological networks can provide specific targets for future experiments. 

There are two different settings under which the link prediction problem is commonly studied.  In the first setting, a snapshot of the network at time $t$, or a sequence of snapshots at times $1,...,t$, is used to 
predict new links that are likely to appear in the near future (at time $t+1$).  In the second setting, the network is treated as static but not fully observed, and the task is to fill in the missing links in such a partially observed network.  These two tasks are related in practice, since a network evolving over time can also be partially observed and a missing link is more likely to emerge in the future.    From the analysis point of view, however, these settings are quite different;  in this paper, we focus on the partially observed setting and do not consider networks evolving over time.  

There are several types of methods for the link prediction problem in the literature.  The first class of methods consists of unsupervised approaches based on various types of node similarities.  These methods assign a similarity score $s(i,j)$ to each pair of nodes $i$ and $j$, and higher similarity scores are assumed to imply higher probabilities of a link.  Similarities can be based either on node attributes or solely on the network structure, such as the number of common neighbors;  the latter are known as structural similarities. Typical choices of structural similarity measures include local indices based on common neighbors, such as the Jaccard index \cite{Liben2007} or the Adamic-Adar index \cite{Adamic2003}, and global indices based on the ensemble of all paths, such as the Katz index \cite{Katz1953} and the Leicht-Holme-Newman Index \cite{Leicht2006}.   Comprehensive reviews of such similarity measures can be found in \cite{Liben2007} and \cite{Lu&Zhou2010}.  

Another class of approaches to link prediction includes supervised learning methods that use both network structures and node attributes. These methods treat link prediction as a binary classification problem, where the responses are $\{1,0\}$ indicating whether there exists a link for a pair, and the predictors are covariates for each pair, which are constructed from node attributes.  A number of popular supervised learning methods have been applied to the link prediction problem. For example, \cite{Ben-Hur2005} and \cite{Bleakley2007} use the support vector machine with pairwise kernels, and \cite{Hasan2006} compares the performance  of several supervised learning methods.     Other supervised methods use probabilistic models for incomplete networks to do link prediction, for example, the hierarchical structure models \cite{Clauset2008}, latent space models \cite{Hoff2002}, latent variable models \cite{Hoff2007,Miller10}, and stochastic relational models \cite{Yu2007}. 

Our approach falls in the supervised learning category, in the sense that we make use of both the node similarities and observed links.   However, one difficulty in treating link prediction as a straightforward classification problem is the lack of certainty about the negative and positive examples.  This is particularly true for negative examples (absent edges).   In biological networks in particular, there may be no certain negative examples at all \cite{Ben-Hur2006}.  For instance, in a protein-protein interaction network, an absent edge may not mean that there is no interaction between the two proteins -- instead, it may indicate that the experiment to test that interaction has not been done, or that it did not have enough sensitivity to detect the interaction.  Positive examples could sometimes also be spurious -- for example, high-throughput experiments can yield a large number of false positive protein-protein interactions \cite{Mering2002}.  Here we propose a new link prediction method that allows for the presence of both false positive and false negative examples.  More formally, we assume that the network we observe is the true network with independent observation errors, i.e., with some true edges missing and other edges recorded erroneously.   The error rates for both kinds of errors are assumed unknown, and in fact cannot be estimated under this framework. However, we can provide rankings of potential links in order of their estimated probabilities, for node pairs with observed links as well as for node pairs with no observed links.  These relative rankings rather than absolute probabilities of edges are sufficient in many applications.  For example, pairs of proteins without observed interactions that rank highly could be given priority in subsequent experiments.  To obtain these rankings, we utilize node covariates when available, and/or network topology based on observed links.

The rest of the paper is organized as follows. In Section \ref{sec:model}, we specify our (rather minimal)  model assumptions for the network and the edge errors. We propose link ranking criteria for both directed and undirected networks in Section \ref{sec:meth}. The algorithms used to optimize these criteria are discussed in Section \ref{sec:alg}. In Section \ref{sec:sim} we compare performance of proposed criteria to other link prediction methods on simulated networks.  In Section \ref{sec:data}, we apply our methods to link prediction in a protein-protein interaction network and a school friendship network.  Section \ref{sec:summary} concludes with a summary and discussion of future directions.

\section{The network model}
\label{sec:model}
A network with $n$ nodes (vertices) can be represented by an $n\times n$ adjacency matrix $A=[A_{ij}]$, where 
  \begin{eqnarray*}
  A_{ij}= \left \{ \begin{tabular}{cc}
               1 & if there is an edge from $i$ to $j$, \\
               0 & otherwise.
               \end{tabular}
      \right .
 \end{eqnarray*}
We will consider the link prediction problem for both undirected and directed networks. Therefore $A$ can be either symmetric (for undirected networks) or asymmetric (for directed networks). 

In our framework, we distinguish between the adjacency matrix of the true underlying network  $A^{True}$, and its observed version $A$.  We assume that each $A^{True}_{ij}$ follows a Bernoulli distribution with $\mathbb{P}(A^{True}_{ij}=1)=P_{ij}$. Given the true network, we assume that the observed network is generated by
\begin{align*}
& \mathbb{P} (A_{ij}=1 |A^{True}_{ij}=1)= \alpha,\quad \mathbb{P} (A_{ij}=0 |A^{True}_{ij}=0)=\beta, 
\end{align*}
where $\alpha$ and $\beta$ are the probabilities of correctly recording a true edge and an absent edge, respectively.   Note that we assume that this probability is constant and does not depend on $i$, $j$, or $P_{ij}$.  
Then we have
\begin{align}
\tilde{P}_{ij}\stackrel{Def}{=}\mathbb{P}(A_{ij}=1)=(\alpha+\beta-1)P_{ij}+(1-\beta).
\label{Ptilde}
\end{align}
If the values of $\alpha$, $\beta$ and $P_{ij}$ were known, then the probabilities of true edges conditional on the observed adjacency matrix could have been estimated as 
\begin{align}
\mathbb{P}(A_{ij}^{True}=1|A_{ij}=1) & = \frac{\alpha P_{ij}}{\tilde{P}_{ij}}, \label{cond1} \\
\mathbb{P}(A_{ij}^{True}=1|A_{ij}=0) & = \frac{(1-\alpha)P_{ij}}{1-\tilde{P}_{ij}}. \label{cond2}
\end{align} 
It is easy to check that both \eqref{cond1} and \eqref{cond2} are monotone increasing functions of $P_{ij}$.   Taking \eqref{Ptilde} into account implies that they are also increasing functions of $\tilde{P}_{ij}$ as long as  $\alpha+\beta>1$.  This gives us a crucial observation: if the goal is to obtain relative rankings of potential links, it is sufficient to estimate $\tilde{P}_{ij}$, and it is not necessary to know $\alpha$, $\beta$ and $P_{ij}$. 


An important special case in this setting is $\beta=1$.  Then all the observed links are true positives, and we only need to provide a ranking for node pairs without observed links.    This can be applied in recommender systems, for example, for recommending possible new friends in a social network.    Another special case is when $\alpha = 1$, which corresponds to all absent edges being true negatives.  This setting can be used to frame the problem of investigating reliability of observed links, for example, in a gene regulatory network inferred from high-throughput gene expression data.   An estimate of $[\tilde{P}_{ij}]$ provides rankings for both these special cases and the general problem, and thus we focus on estimating $\tilde{P}_{ij}$ for the rest of the paper.

\section{Link prediction criteria}\label{sec:meth}

In this section, we propose criteria for estimating the probabilities of edges in the observed network, $\tilde P_{ij}$, for both directed and undirected networks. The criteria rely on a symmetric matrix $W=[W_{ii'}]$ with $0 \leq W_{ii'} \leq 1$, which describes the similarity between nodes $i$ and $i'$. The similarity matrix $W$ can be obtained from different sources, including node information, network topology, or a combination of the two.   We will discuss choices of $W$ later in this section. 
\subsection{Link prediction for directed networks}
First we consider directed networks.
\begin{figure}[h!]
\begin{center}
\includegraphics[width=5.0cm]{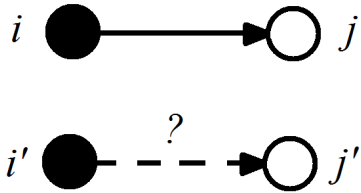}
\caption{Pair similarity for directed networks}
\label{fig:directed-intuition}
\end{center}
\end{figure}
The key assumption we make is that if two pairs of nodes are similar to each other, the probability of links within these two pairs are also similar.  Specifically, in Figure \ref{fig:directed-intuition}, $P_{ij}$ and $P_{i'j'}$ are assumed close in value if node $i$ is similar to node $i'$ and node $j$ is similar to node $j'$.
For directed networks, we measure similarity of node pairs $(i,i')$ and $(j,j')$ by the product $W_{ii'}W_{jj'}$ (see Figure \ref{fig:directed-intuition}), which implies two pairs are similar only if both pairs of endpoints are similar.   This assumption should not to be confused with a different assumption made by many unsupervised link prediction methods,  which assume that a link is more likely to exist between similar nodes, applicable to networks with assortative mixing.   Assortative networks are common -- a typical example is a social network, where people commonly tend to be friends with those of similar age, income level, race, etc. However, there are also networks with disassortative mixing, in which the assumption that similar pairs are more likely to be connected is no longer valid -- for example, predators do not typically feed on each other in a food web. Our assumption, in contrast, is equally plausible for both assortative and disassortative networks, as well as more general settings, as it does not assume anything about the relationship between $P_{ij}$ and $W_{ij}$.

Motivated by this assumption of similar probabilities of links for similar node pairs, we propose to estimate $\tilde P_{ij} = E(A_{ij})$ by 
\begin{align}\label{directed-pu}
\hat f =  \argmin_{f} \frac{1}{n^2}\sum_{ij}^n (A_{ij}-f_{ij})^2+ \frac{\lambda }{n^4} \sum_{ii'jj'}^n W_{ii'}W_{jj'} (f_{ij}-f_{i'j'})^2,
\end{align} 
where $f$ is a real-valued $n \times n$ matrix, and $\lambda$ is a tuning parameter.  The first term is the usual squared error loss connecting the parameters with the observed network. The minimizer of its population version, i.e., $\mathbb{E}(A_{ij}-f_{ij})^2$ is $\tilde{P}_{ij}$.  The second term enforces our key assumption,  penalizing the difference between $f_{ij}$ and $f_{i'j'}$ more if two node pairs $(i,i')$ and $(j,j')$ are similar. 
The choice of the squared error loss is not crucial, and other commonly used loss functions could be considered instead, for example, the hinge loss or the negative log-likelihood.  The main reason for choosing the squared error loss is computational efficiency, since it makes \eqref{directed-pu} a quadratic problem;  see more on this details in Section \ref{sec:alg}.

In some applications, we may have additional information about true positive and negative examples, i.e., some $A_{ij}$'s may be known to be true 1's and true 0's, while others may be uncertain.  This could happen, for example, when validation experiments have been conducted on a subset of a gene or protein network inferred from expression data.   If such information is available, it makes sense to use it, and we can then modify  criterion \eqref{directed-pu} as follows: 
\begin{align}\label{directed-semi}
\argmin_{f} \frac{1}{\sum_{ij}^n E_{ij}}\sum_{ij}^n E_{ij}(A_{ij}-f_{ij})^2+ \frac{\lambda }{n^4} \sum_{ii'jj'}^n W_{ii'}W_{jj'} (f_{ij}-f_{i'j'})^2,
\end{align}
where $E_{ij}=1$ if it is known that $A_{ij} = A^{True}_{ij}$, and 0 otherwise. This is similar to a semi-supervised criterion proposed in \cite{Kashima09}.  However, \cite{Kashima09} did not consider the uncertainty in positive and negative examples, nor did they consider the undirected case which we discuss next. Since \eqref{directed-semi} only involves a partial sum of the loss function terms, we will refer to \eqref{directed-semi} as the partial-sum criterion and \eqref{directed-pu} as the full-sum criterion for the rest of the paper. 

\subsection{Link prediction for undirected networks}
\begin{figure}[h!]
 \begin{center}
\twoImages{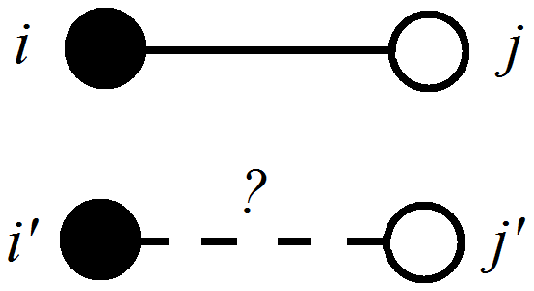}{5cm}{}
{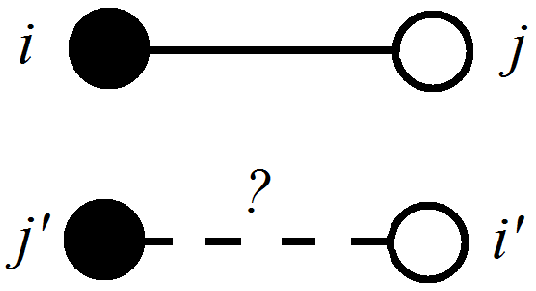}{5cm}{}
\end{center}
\caption{Pair similarity for undirected networks}
\label{fig:undirected-intuition}
\end{figure}
For undirected networks, our key assumption that $P_{ij}$ and $P_{i'j'}$ are close if two pairs $(i,i')$ and $(j,j')$ are similar needs to take into account that the direction no longer matters;  thus the pairs are similar if either $i$ is similar to $i'$ and $j$ is similar to $j'$, or if $i$ is similar to $j'$ and $j$ is similar to $i'$ (see Figure \ref{fig:undirected-intuition}.   Thus we need a new pair similarity measure that combines $W_{ii'}W_{jj'}$ and $W_{ij'}W_{ji'}$.  There are multiple options; for example, two natural combinations are
\begin{align*}
S_1=W_{ii'}W_{jj'}+W_{ij'}W_{ji'} \mbox{ and } S_2=\max(W_{ii'}W_{jj'},W_{ij'}W_{ji'}). 
\end{align*}
Empirically, we found that $S_2$ performs better than $S_1$ for a range of real and simulated networks.  The reason for this can be easily illustrated on the stochastic block model.   The stochastic block model is a commonly used model for networks with communities, where the probability of a link only depends on the community labels of its two endpoints. Specifically, given community labels $c = \{c_1, \dots, c_n\}$, $A^{True}_{ij}$'s are independent Bernoulli random variables with  
\begin{equation}
P_{ij}  = S_{c_i c_j}, \label{bm}
\end{equation}
where $S=[S_{ab}]$ is a $K \times K$ symmetric matrix, and $K$ is the number of communities in the network.   Suppose we have the best similarity measure we can possibly hope to have based on the truth,   $W_{ij}=I(c_i=c_j)$, where $I$ is the indicator function.   In that case, \eqref{bm} implies $P_{ij}=P_{i'j'}$ if $\max(W_{ii'}W_{jj'},W_{ij'}W_{ji'})=1$, whereas the sum of the weights would be misleading. 

Using $S_2$ as the measure of pair similarity, we propose estimating $\tilde P_{ij}$ for undirected networks by 
\begin{eqnarray}\label{undirected-pu}
\argmin_{f} && \frac{1}{n^2}\sum_{i<j}^n (A_{ij}-f_{ij})^2 + \\
&& \frac{\lambda }{n^4} \sum_{i<j,i'<j'}^n \max(W_{ii'}W_{jj'},W_{ij'}W_{ji'}) (f_{ij}-f_{i'j'})^2. \nonumber
\end{eqnarray} 
Similarly to the directed case, if we have information about true positive and negative examples, we can use a partial-sum criterion
\begin{eqnarray}\label{undirected-semi}
\argmin_{f} && \frac{1}{\sum_{i<j}^n E_{ij}}\sum_{i<j}^n E_{ij}(A_{ij}-f_{ij})^2 + \\
&& \frac{\lambda }{n^4} \sum_{i<j,i'<j'}^n \max(W_{ii'}W_{jj'},W_{ij'}W_{ji'}) (f_{ij}-f_{i'j'})^2, \nonumber
\end{eqnarray} 
where $E_{ij}=1$ if it is known that $A_{ij} = A^{True}_{ij}$, otherwise $E_{ij}=0$.

\subsection{Node similarity measures}
The last component we need to specify is the node similarity matrix $W$.    One typical situation is when we have reasons to believe that the external node covariates are related to the structure of the network, in which case it is natural to use covariate information to construct $W_{ii'}$.  Though more complicated formats do exist, node covariates are typically represented by an $n \times p$ matrix $X$  where $X_{ik}$ is the value of variable $k$ on node $i$.  Then $W_{ii'}$ can be taken to be some similarity measure between the $i$-th and $i'$-th rows of $X$. For example, if $X$ contains only numerical variables and has been standardized, we can use the exponential decay kernel, 
$$W_{ii'}= \exp \left \{ -\frac{\| X_i-X_{i'} \|^2}{\sigma^2} \right \},$$
where $\| \cdot \|$ is the Euclidean vector norm. 

When node covariates are not available, node similarity $W_{ii'}$ is usually  obtained from the topology of the observed network $A$, i.e., $W_{ii'}$ is large if $i$ and $i'$ have a similar pattern of connections with other nodes.  For undirected networks, a simple choice of $W_{ii'}$ could be
\begin{align}
W_{ii'}= \frac{ |\{k: A_{ik}=A_{i'k}\} |} {n} \ , \label{topo1}
\end{align} 
where $| \cdots |$ denotes cardinality of a set.   
This particular measure turns out to be not very useful:  since most real networks are sparse, most entries of any $k$-th column will be 0, and thus most of $W_{ii'}$'s would be large. A more informative measure is the Jaccard index \cite{Liben2007}, 
\begin{align}
W_{ii'}=\frac{ |N(i) \cap N(i')|} {|N(i) \cup N(i') |}, \label{topo2}
\end{align}
where $N(i) = \{k: A_{ik} = 1 \}$ is the set of neighbors of node $i$.   

The directed networks case is similar, except we need to count the in and the out links separately.  The formulas corresponding to \eqref{topo1} and \eqref{topo2} become
\begin{align*}
W_{ii'} & = \frac{|\{k: A_{ik}=A_{i'k} \}|}{2n}+\frac{|\{k: A_{ki}=A_{ki'} \}|}{2n}, \\
W_{ii'} & = \frac{ |N_1(i) \cap N_1(i')| }{2 |N_1(i) \cup N_1(i') |}+\frac{ |N_2(i) \cap N_2(i')|}{2|N_2(i) \cup N_2(i') |},
\end{align*}
where $N_1(i)=\{k: A_{ik}=1 \} $ and $N_2(i)=\{k:  A_{ki}=1 \} $.

\section{Optimization algorithms}\label{sec:alg}
The proposed link prediction criteria are convex and quadratic in parameters, and thus optimization is fairly straightforward.   The obvious approach is to treat the matrix $f$ as a long vector with $n^2$ elements (or  $n(n-1)/2$ in the undirected case), and solve the linear system obtained by taking the first derivative of any  criterion above with respect to this vector.   However, solving a system of linear equations could be challenging for large-scale problems \cite{Boyd04};  the number of parameters here is $O(n^2)$, and so the linear system requires $O(n^4)$ memory.  However, if $W$ is sparse, or sparsified by applying thresholding or some other similar method, then solving the linear system is the efficient choice.

If the $W$ matrix is not sparse, an iterative algorithm with sequential updates that only requires $O(n^2)$ memory would be a better choice than solving the linear system.   We propose an iterative algorithm following the idea of block coordinate descent \cite{Hildreth57, Warga63}.   A block coordinate descent algorithm partitions the coordinates
into blocks and iteratively optimizes the criterion with respect to each block while holding the other blocks fixed. 

First, we derive the update equations for directed networks.  Note \eqref{directed-pu} and \eqref{directed-semi} can be written in the general form
\begin{equation} \label{directed-general}
 Q = \frac{1}{n^2}\sum_{ij}^n V_{ij}(A_{ij}-f_{ij})^2+ \frac{\lambda }{n^4} \sum_{ii'jj'}^n W_{ii'}W_{jj'} (f_{ij}-f_{i'j'})^2, \end{equation} 
where $V_{ij} \equiv 1$ for \eqref{directed-pu} and $V_{ij}=E_{ij}$ for \eqref{directed-semi}. For any matrix $M$, let $M_{i \cdot}$ be the $i$th row of $M$. We treat $f_{i \cdot }$ as a block, and update $f_{i \cdot}$ iteratively. Define $V_i=\F{diag}(V_{i \cdot})$. Then
\begin{align}\label{term1}
\sum_{ij} V_{ij}(f_{ij}-A_{ij})^2 & =\sum_i (f_{i\cdot}-A_{i\cdot})^T V_i(f_{i\cdot}-A_{i\cdot}).
\end{align}
Let $D$ be an $n \times n$ diagonal matrix with $D_{ii}=\sum_{j} W_{ij}$. Then
\begin{align}
\sum_{jj'} W_{jj'} (f_{ij}-f_{i'j'})^2 & = f_{i\cdot}^T D f_{i\cdot}-2f_{i\cdot}^T W f_{i'\cdot}+ f_{i'\cdot}^T D f_{i'\cdot} \label{term2}
\end{align}
Plugging \eqref{term1} and \eqref{term2} into \eqref{directed-general}, and taking the first derivative of $Q$ with respect to $f_{i \cdot}$, we obtain
\begin{eqnarray}
\frac{\partial Q}{\partial f_{i\cdot}} &=& \frac{2}{n^2} (f_{i\cdot}-A_{i\cdot}) V_i + \\
&& \lambda \frac{4}{n^4}\left[W_{ii}(Df_{i\cdot}-Wf_{i\cdot})+\sum_{i'\neq i} W_{ii'}(Df_{i\cdot}-Wf_{i'\cdot}) \right]. \nonumber
\end{eqnarray}
Solving $ \frac{\partial Q}{\partial f_{i\cdot}} =0$ with respect to $f_{i \cdot}$, we obtain the updating formula  
\begin{align}\label{update1}
f^{(t+1)}_{i\cdot} & \leftarrow \left( n^2 V_i +2\lambda \sum_{i'} W_{ii'}D-2\lambda W_{ii}W \right)^{-1} \left (n^2 A_{i\cdot}V_i+2\lambda \sum_{i' \neq i} W_{ii'} W f_{i'\cdot}^{(t)} \right), 
\end{align}
where $f_{i\cdot}^{(t)}$ is the value of $f_{i\cdot}$ at iteration $t$.

This update is fast to compute but its derivation relies on the product form of $W_{ii'}$ and $W_{jj'}$, and thus is not directly applicable in the undirected case, where $S_2$ is used as the similarity measure.    However, we can still approximate $S_2$ with a product, using the fact that for $x \geq 0, y \geq 0$, $\lim_{q \rightarrow \infty} \sqrt [q] {x^q+y^q}= \max(x,y)$. Thus, for sufficiently large $q$, we have 
\begin{align}
[\max(W_{ii'}W_{jj'},W_{ij'}W_{ji'})]^q \approx (W_{ii'}W_{jj'})^q +(W_{ij'}W_{ji'})^q. \label{appx}
\end{align}
Further, $W^q$ is a monotone transformation of $W$ and can also serve as a similarity measure.   Based on \eqref{appx}, we propose to substitute the following approximate criterion for undirected networks,
\begin{eqnarray} \label{undirected-general}
 Q &=& \frac{1}{n^2}\sum_{i<j}^n V_{ij} (A_{ij}-f_{ij})^2 + \\
 && \frac{\lambda }{n^4} \sum_{i<j,i'<j'}^n ((W_{ii'}W_{jj'})^q +(W_{ij'}W_{ji'})^q ) (f_{ij}-f_{i'j'})^2, \nonumber
\end{eqnarray}
where $V_{ij}\equiv 1$ for the full sum criterion and $V_{ij}=E_{ij}$ for the partial sum criterion. By symmetry, 
\begin{align*}
  & \sum_{i<j,i'<j'}^n ((W_{ii'}W_{jj'})^q +(W_{ij'}W_{ji'})^q ) (f_{ij}-f_{i'j'})^2 \\
= & \frac{1}{2} \sum_{i \neq j ,i' \neq j'}^n W_{ii'}^q  W_{jj'}^q (f_{ij}-f_{i'j'})^2 .
\end{align*} 
This is now in the same form as \eqref{directed-general}, with each term in the sum containing a product of  $W_{ii'}$ and $W_{jj'}$, and therefore \eqref{undirected-general} can be solved by block coordinate descent with an analogous updating equation as that in the directed network case.  

In practice, we found that when $W$ is sparse or truncated to be sparse, solving the linear system can be much faster than the block coordinate descent method; however, when $W$ is dense and the number of nodes is reasonably large, the block coordinate descent method dominates directly solving linear equations.

\section{Simulation studies} \label{sec:sim}
\begin{figure}[htb!]
\twoImages{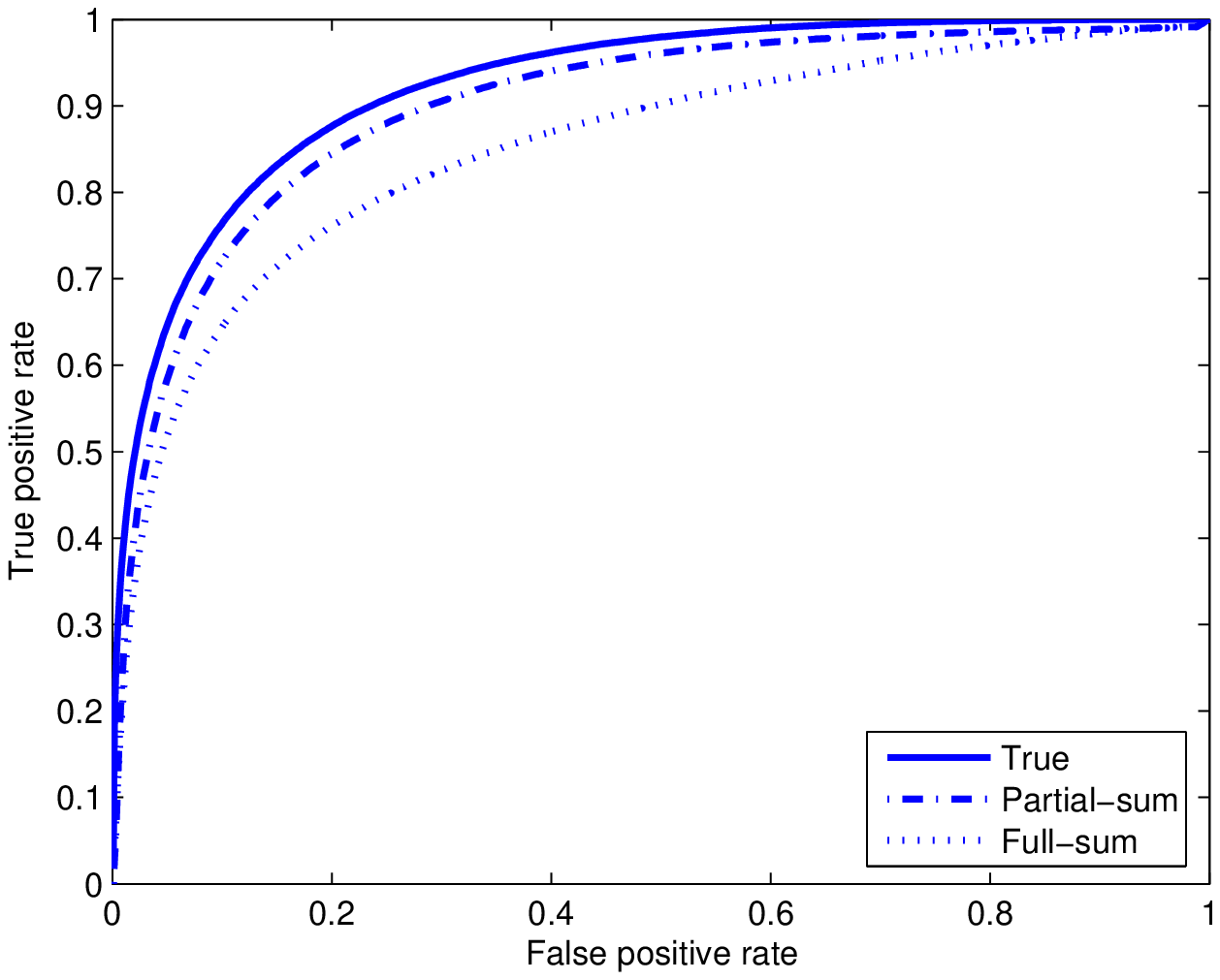}{7cm}{$(a) \quad d =500$}
{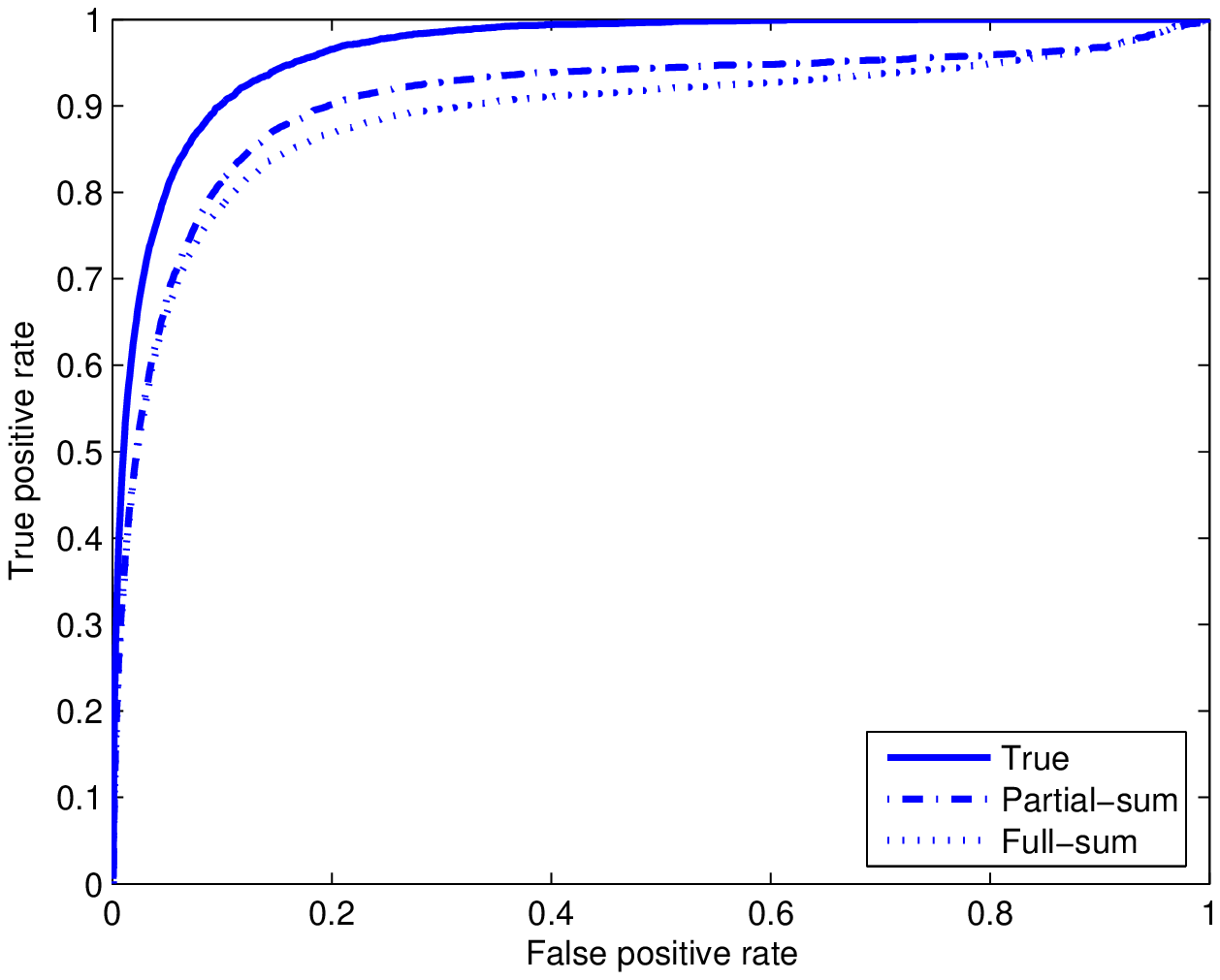}{7cm}{$(a') \quad d =13 $}
\twoImages{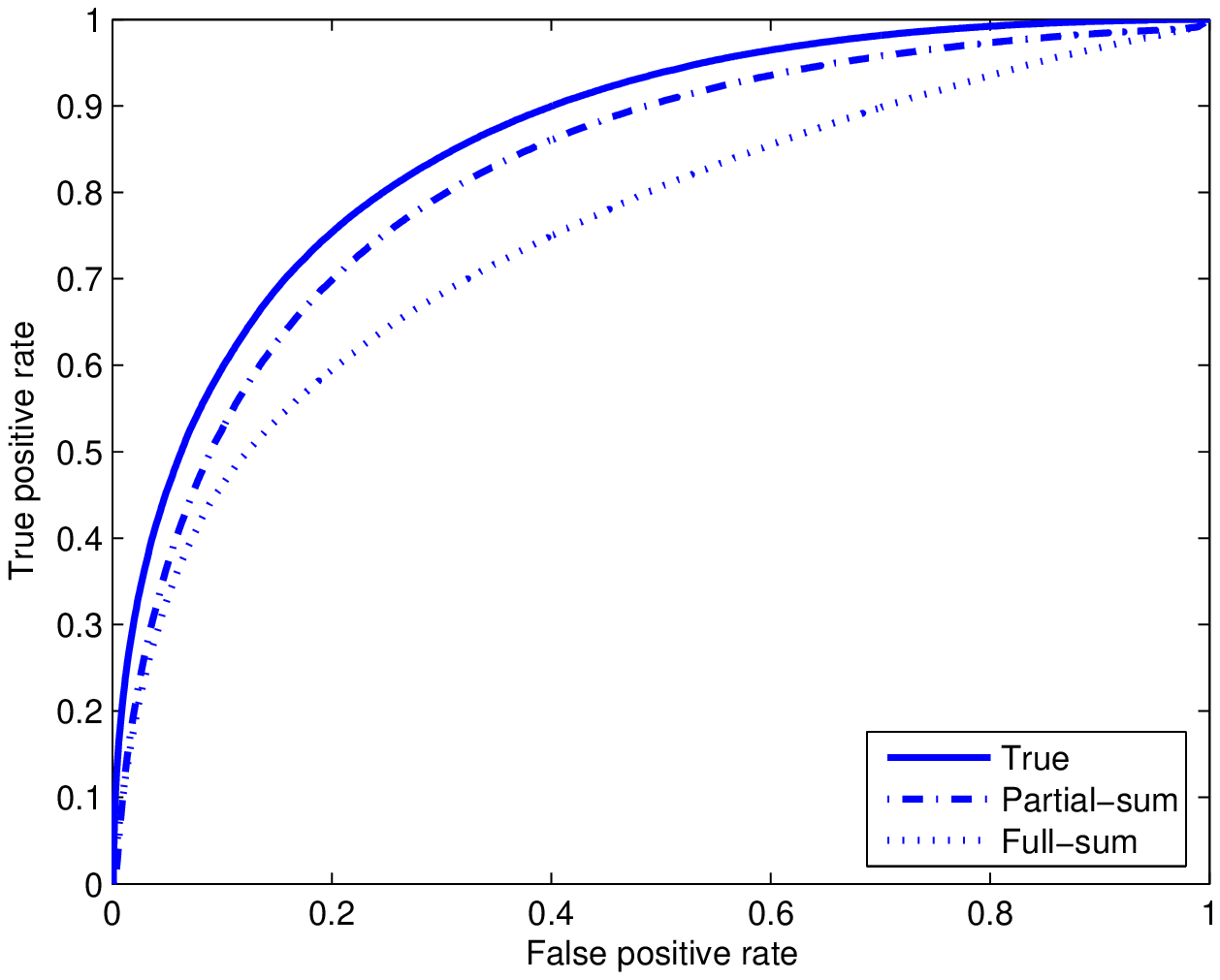}{7cm}{$(b) \quad d = 500$}
{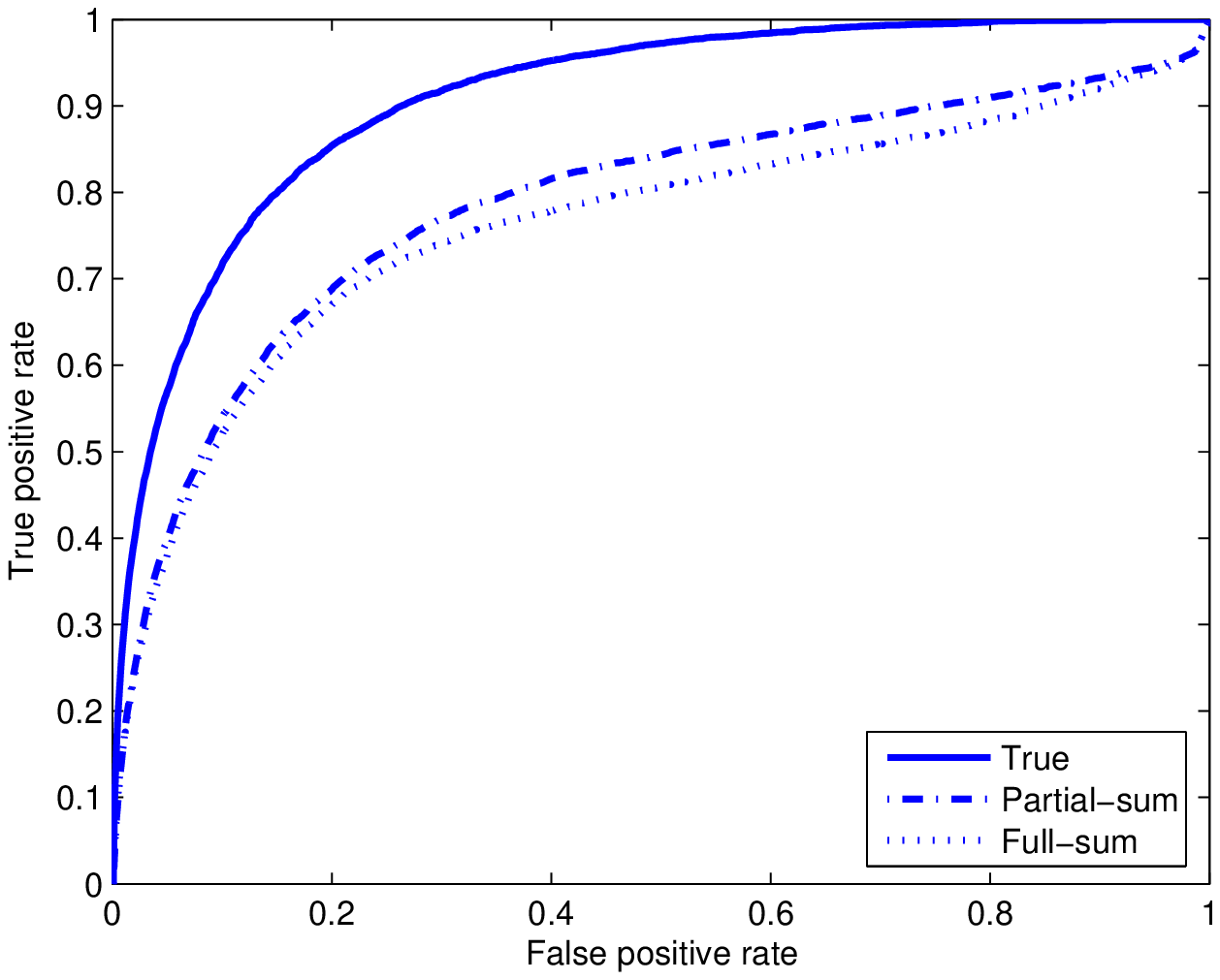}{7cm}{$(b') \quad d =15 $}
\caption{ROC curves for directed networks. $d$ is the average degree over 20 replicates.}
\label{fig:sim1}
\end{figure}

\begin{figure}[htb!]
\twoImages{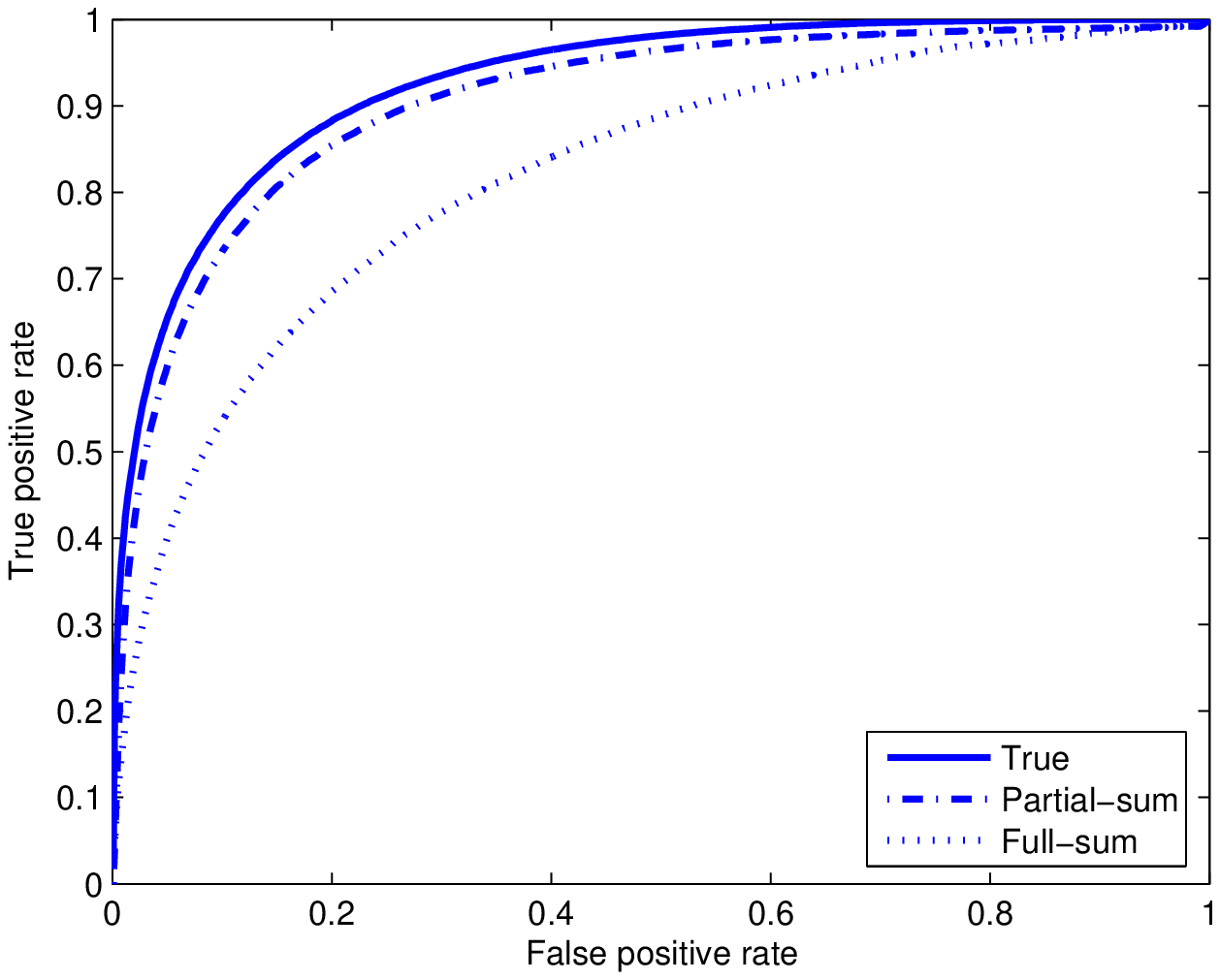}{7cm}{$(c)  \quad  d = 500$}
{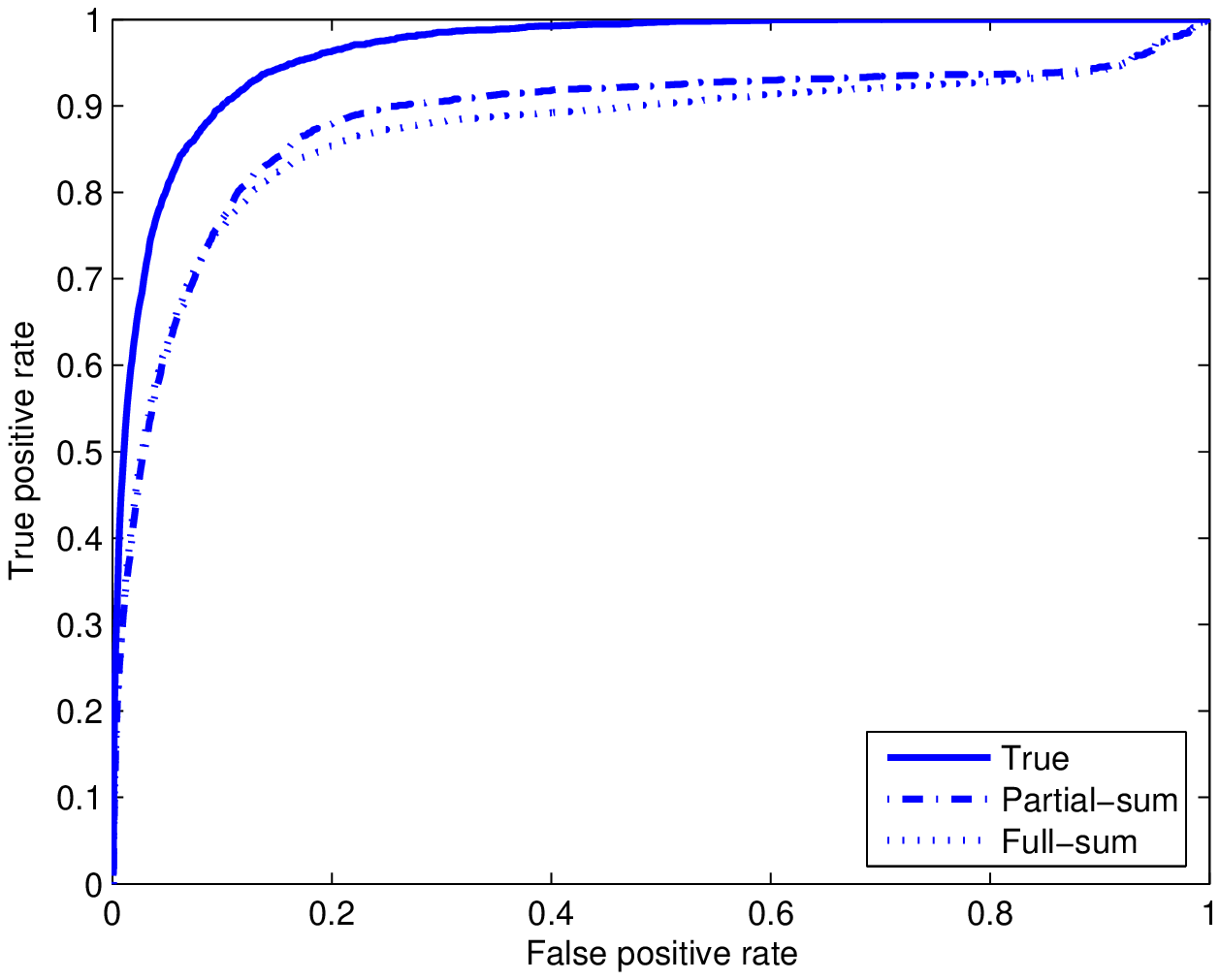}{7cm}{$(c') \quad  d = 13$}
\twoImages{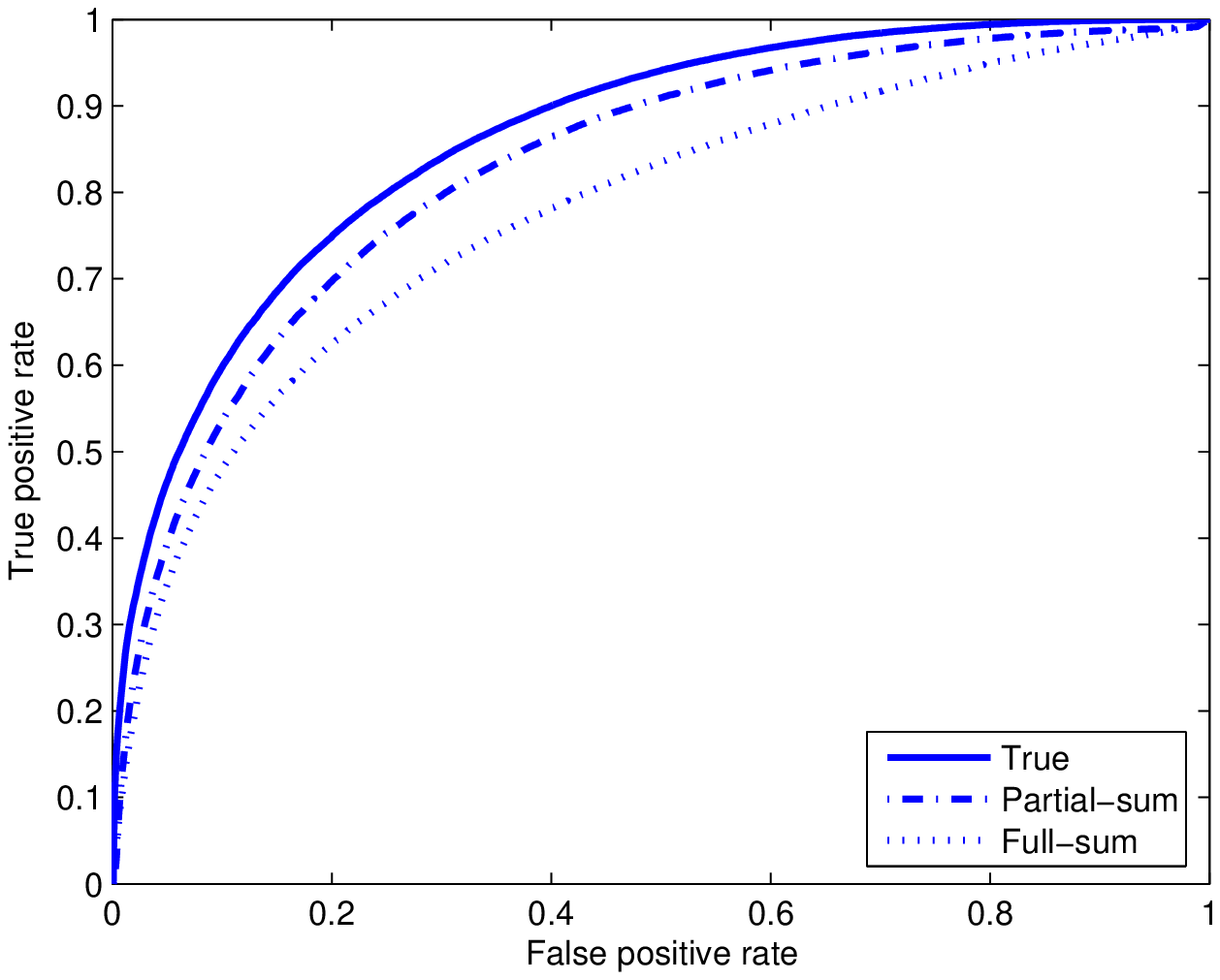}{7cm}{$ (d) \quad  d =500$}
{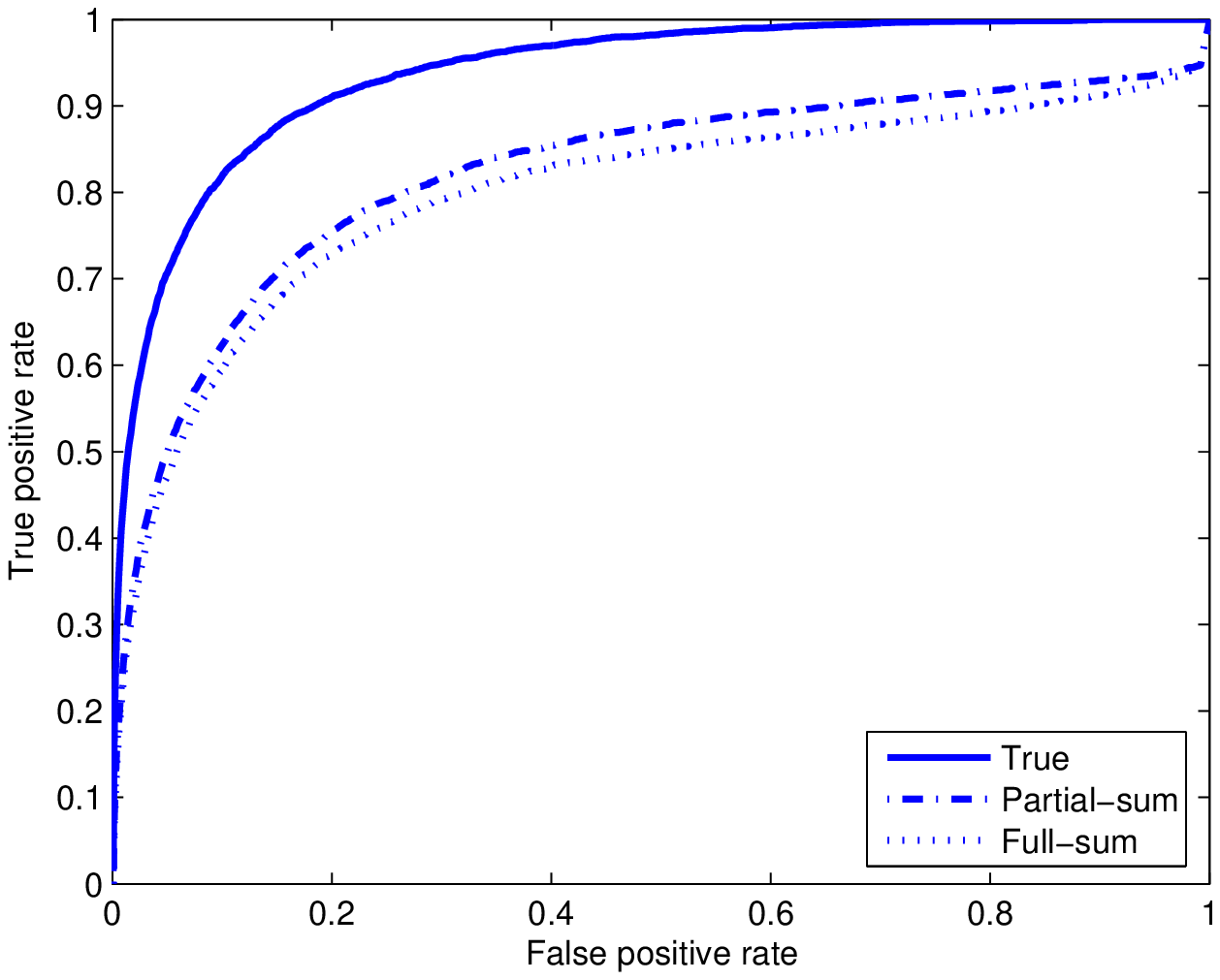}{7cm}{$(d') \quad  d =20$}
\caption{ROC curves for undirected networks. $d$ is the average degree over 20 replicates.}
\label{fig:sim2}
\end{figure}

In this section, we test performance of our link prediction methods on simulated networks.  In all cases, each network consists of $n=1000$ nodes, and node $i$'s covariates $X_i$ are independently generated from a multivariate normal distribution $N_p(0,I_p)$ with $p=5$. 
Each $A_{ij}^{True}$ is generated independently, with $\mbox{logit} P_{ij} = f(X_i, X_j)$.  We consider the following functions $f(X_i, X_j)$:
\begin{align*}
& (a) \quad \sum_{k} (X_{ik}-X_{jk}),  & (a')  & \quad \sum_{k} (X_{ik}-X_{jk})-8, \\
& (b) \quad 2X_i^T X_j/\|X_j\|,  & (b') & \quad X_i^T X_j/\|X_j\|-6, \\
& (c) \quad \sum_{k} (X_{ik}+X_{jk}), & (c') & \quad\sum_{k} (X_{ik}+X_{jk})-8, \\
& (d) \quad X_i^T X_j, & (d') & \quad X_i^T X_j-6. 
\end{align*}
The right hand column gives sparser versions of functions in the left hand column (subtracting a constant within the logit link functions lowers the overall degree), which we use to compare dense and sparse networks (the  average degrees of all these networks are reported in Figures \ref{fig:sim1} and \ref{fig:sim2}).  Functions (a) and (b) are asymmetric in $X_i$ and $X_j$, giving directed networks, while (c) and (d) are symmetric functions corresponding to undirected networks. Further, $(a)$ and $(c)$ are linear functions; $(b)$ is the projection model proposed in \cite{Hoff2002}, under which the link probability is determined by the projection of $X_i$ onto the direction of $X_j$, and $(d)$ is an undirected version of the projection model. 

We also generate indicators $E_{ij}$'s as independent Bernoulli variables taking values 1 and 0 with equal probability, and set $A_{ij}=  E_{ij} A^{True}_{ij}$. This setup corresponds to the ``partially observed'' network of the title, where all the observed edges are true but the missing edges may or may not be true 0s.     

Since we have node covariates affecting the probabilities of links in this case, we define the similarity matrix $W$ by 
$$W_{ii'}= \exp \left \{ -\frac{\| X_i-X_{i'} \|^2}{\sigma^2} \right \},$$
where we choose $\sigma=\frac{1}{4} \mbox{median} \{ \| X_i-X_{i'} \|, i=1,...,n, i'=1,...,n \} $. After truncating $W$ at 0.1, we optimize all criteria by solving linear equations, with $\lambda$ chosen by 5-fold cross validation. 

The performance of link prediction is evaluated on the ``test'' set $\{(i,j): E_{ij}=0\}$.  We report ROC curves, which only depend on the rankings of the estimates $\hat{f}_{ij}$ rather than their numerical values.  Specifically, let $R_{ij}$ be the ranking of $\hat{f}_{ij}$ on the test set in descending order. For any integer $k$, we define false positives as pairs $(i,j)$ ranked within top $k$ but without links in the true network ($A_{ij}^{True} = 0$), and true positives as pairs ranked within top $k$ with $A_{ij}^{True}=1$. Then the true positive rate (TPR) and the false positive rate (FPR) are defined by 
\begin{align*}
\F{TPR}(k)= \frac{|\{(i,j): E_{ij}=0, R_{ij} \leq k, A_{ij}^{True}=1 \}| }{| \{ (i,j): E_{ij}=0, A_{ij}^{True}=1 \} |}, \\
\F{FPR}(k)= \frac{| \{ (i,j): E_{ij}=0, R_{ij} \leq k, A_{ij}^{True}=0 \} |}{| \{ (i,j): E_{ij}=0, A_{ij}^{True}=0 \} |}.
\end{align*}
The ROC curves showing the false positive rate vs.\ the true positive rate over a range of $k$ values are shown in Figures \ref{fig:sim1} (directed networks) and \ref{fig:sim2} (undirected networks).  Each curve is the average of 20 replicates.   We also show the ROC curve constructed from true $P_{ij}$'s as a benchmark..  

Overall, both the full sum and the partial sum criteria perform well.  There is little difference between directed network models and their undirected versions. As expected, the partial sum criterion always gives better results since it has more information and only uses the true positive and negative examples for training. But its performance is quite comparable to the completely unsupervised full sum criterion, except perhaps for model $(c)$.  The gaps between the unsupervised full sum criterion and semi-supervised partial sum criterion become smaller for sparse networks, as the false negatives in the full sum are only a small proportion of the large number of true negatives in a sparse network.  The ROC curve obtained from the true model in sparse networks is better than in the corresponding dense networks;  this seemingly counter-intuitive finding is also explained by the large number of 0s in sparse networks.   However, gaps between both our link prediction methods  and the true model are larger in all the sparse networks than in their dense counterparts. This confirms the observation that a small number of positive examples in sparse networks makes the link prediction problem challenging.

\section{Applications} \label{sec:data}
\subsection{The protein-protein interaction network}
Our first application is to an undirected network containing yeast protein-protein interactions from \cite{Mering2002}.  This network was edited to contain only highly reliable interactions supported by multiple experiments \cite{Bleakley2007}, resulting in 984 protein nodes and 2438 edges, with the average node degree about 5.  We take this verified network to be the true underlying network $A^{True}$. \cite{Bleakley2007} also constructed a matrix measuring similarities between proteins based on gene expression, protein localization, phylogenetic profiles and yeast two-hybrid data, which we use as the node similarity matrix $W$ for link prediction. 

\begin{figure}[h!]
\threeImages{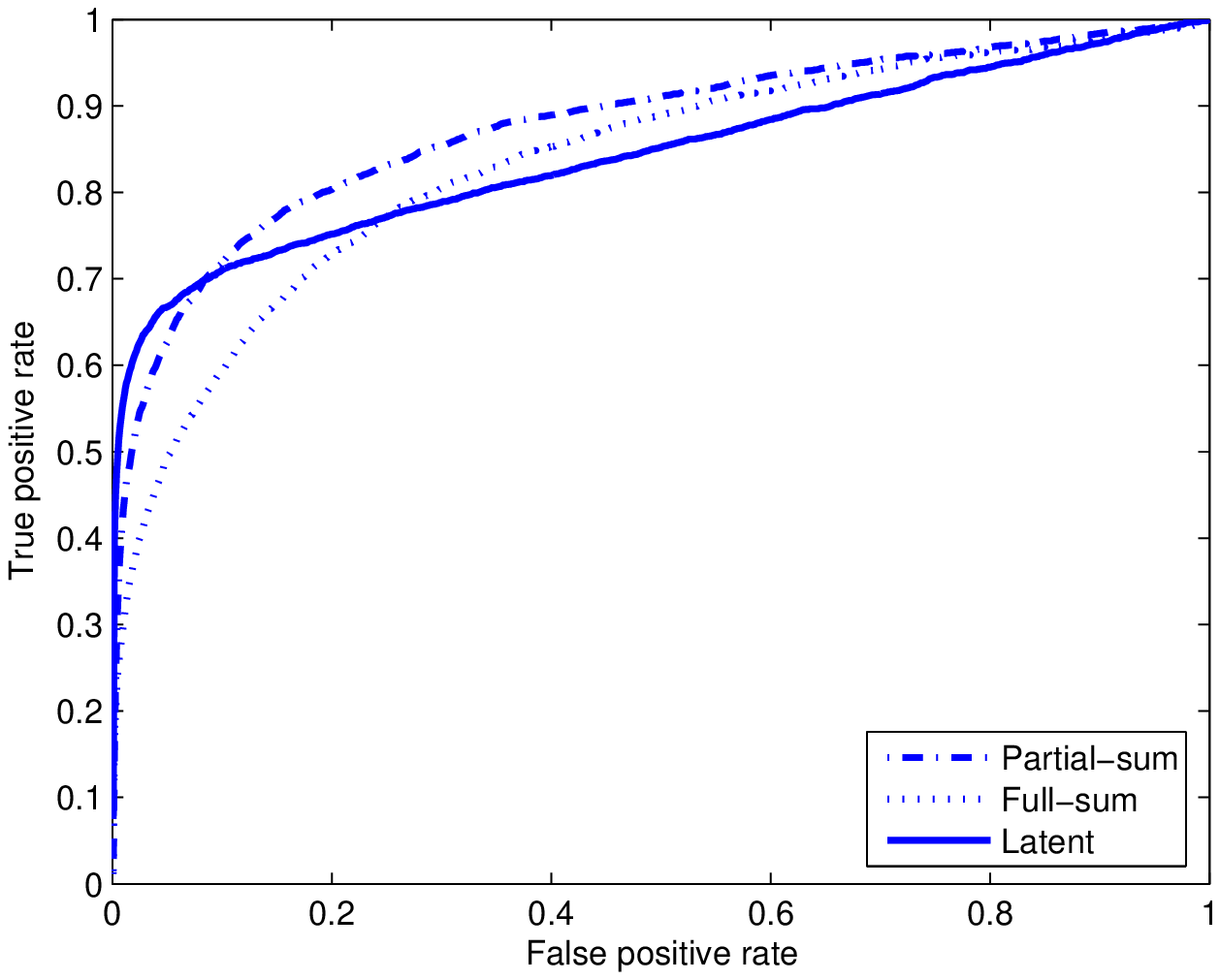}{5cm}{(a) $\alpha=0.8$}{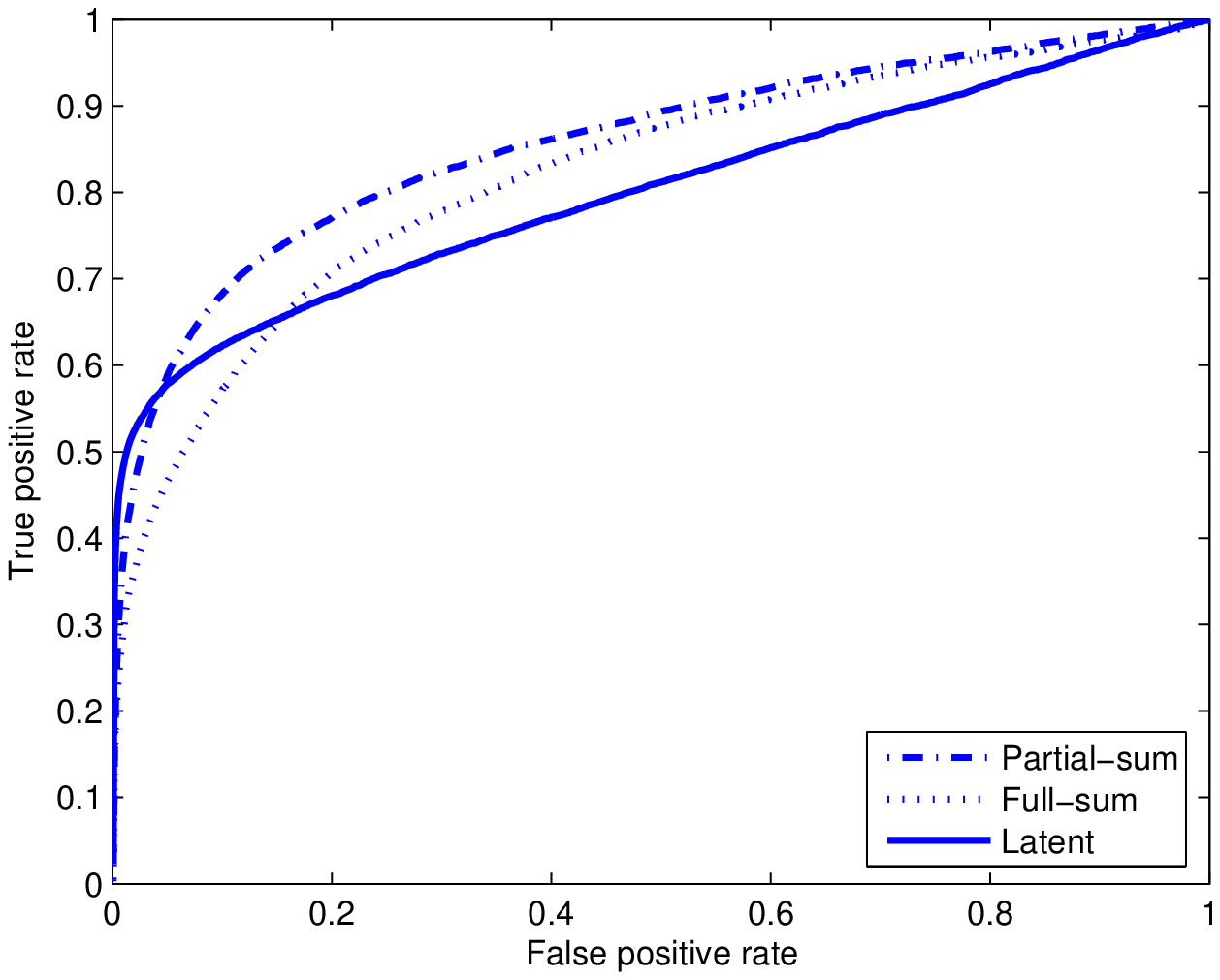}{5cm}{(b) $\alpha=0.5$ }{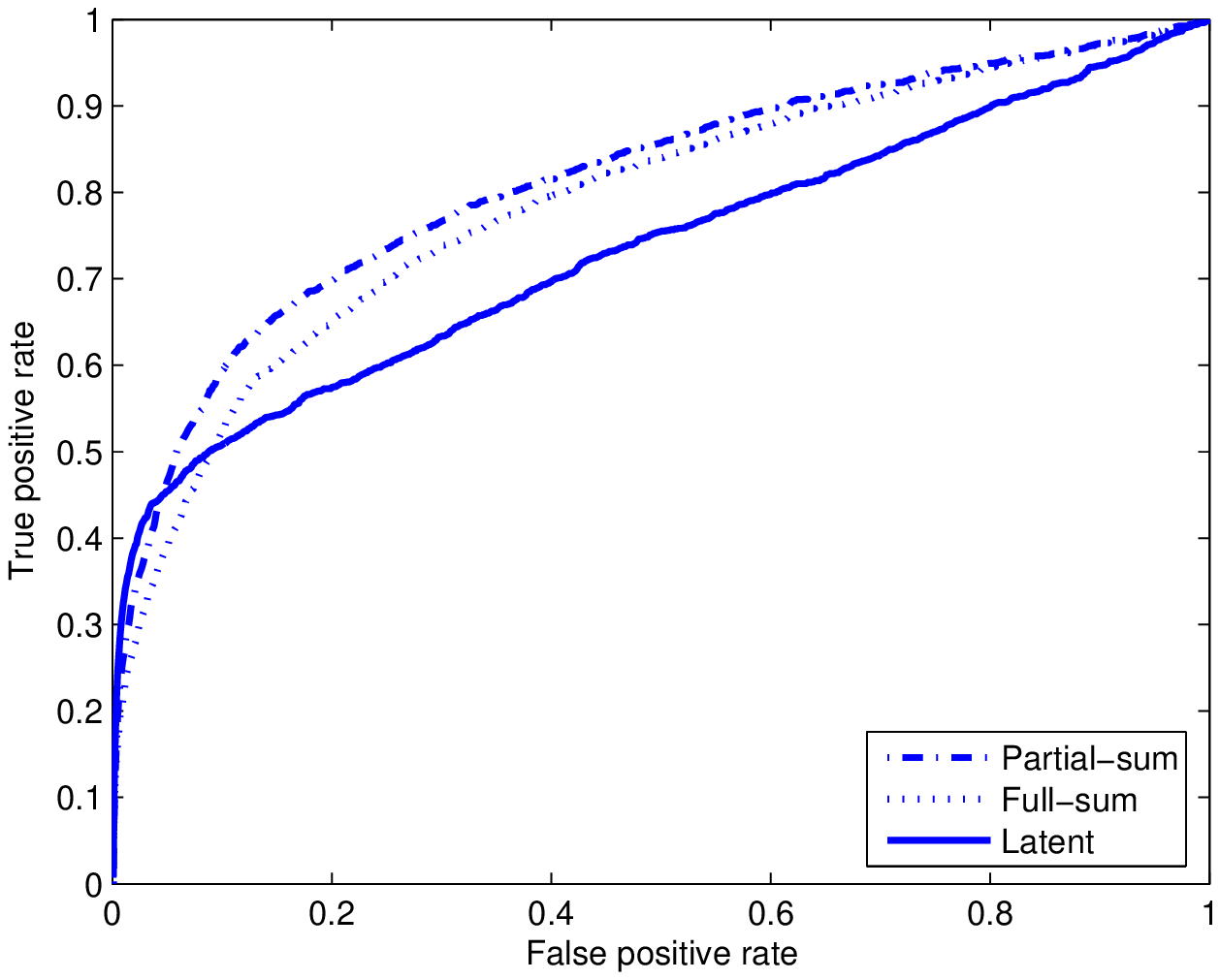}{5cm}{(c) $\alpha=0.2$ }
\caption{ROC curves for the protein-protein interaction network. }
\label{fig:PPI}
\end{figure}  

Here, we compare  the full sum criterion \eqref{undirected-pu}, the partial sum criterion \eqref{undirected-semi}, and the latent variable model proposed by \cite{Hoff2007}. To test prediction, we generate 
indicators $E_{ij}$'s as independent Bernoulli variables taking value 1 with probability $\alpha$, and set  $A_{ij}=E_{ij} A^{True}_{ij}$.  We consider three different values of $\alpha$, $\alpha=0.2,0.5,0.8$, corresponding to different amounts of available information.  
 
We use the block coordinate descent algorithm proposed in Section \ref{sec:alg} to approximately optimize \eqref{undirected-pu} and \eqref{undirected-semi}, with $q=10$ and $\lambda$ chosen by cross-validation. The latent variable model depends on a tuning parameter $K$, the dimension of the latent space. We fix $K=5$ since larger values of $K$ do not significantly change the performance in this example.   We again use ROC curves to evaluate the link prediction performance on the set $\{(i,j): E_{ij}=0\}$. Each ROC curve in Figure \ref{fig:PPI} is the average of 10 random realizations of $E_{ij}$'s.  
 
The semi-supervised criterion always performs better than the unsupervised criterion, as it should.  Further, the semi-supervised criterion almost always outperforms the latent variable model, except for very small values of the false positive rate, and the fully unsupervised criterion also starts to outperform the latent variable model  as the false positive rate increases.  The latent variable model is also more sensitive to the sampling rate $\alpha$, with performance deteriorating for $\alpha = 0.2$.  This is because the model relies heavily on the structure of the network, and a low sampling rate may substantially distort the overall network topology. On the other hand, we use the node similarity matrix $W$ which depends only on the features of the proteins, and is thus unaffected by the sampling rate.  

\subsection{The school friendship network}

\begin{figure}[h!]
\threeImages{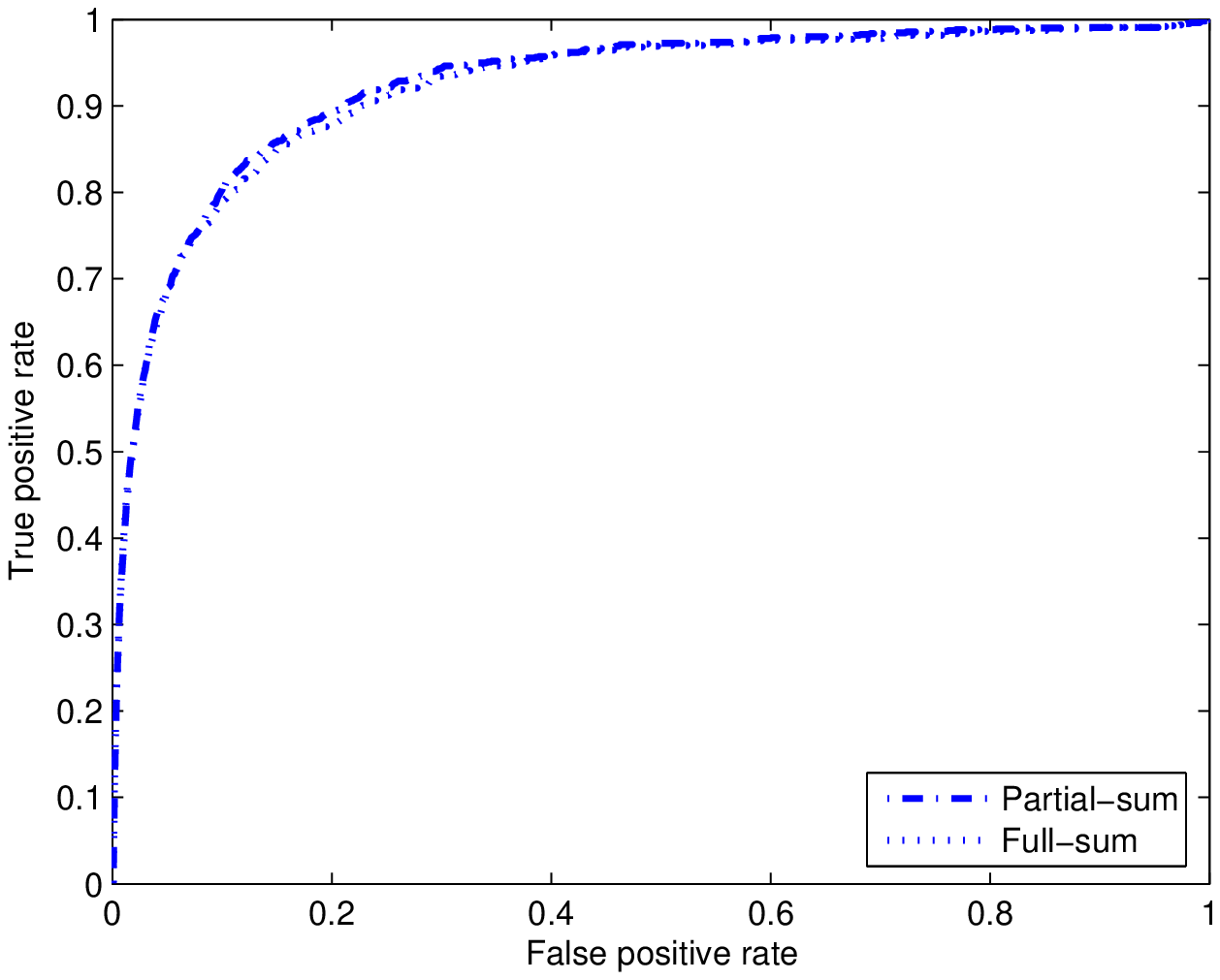}{5cm}{(a) $\alpha=0.8$}{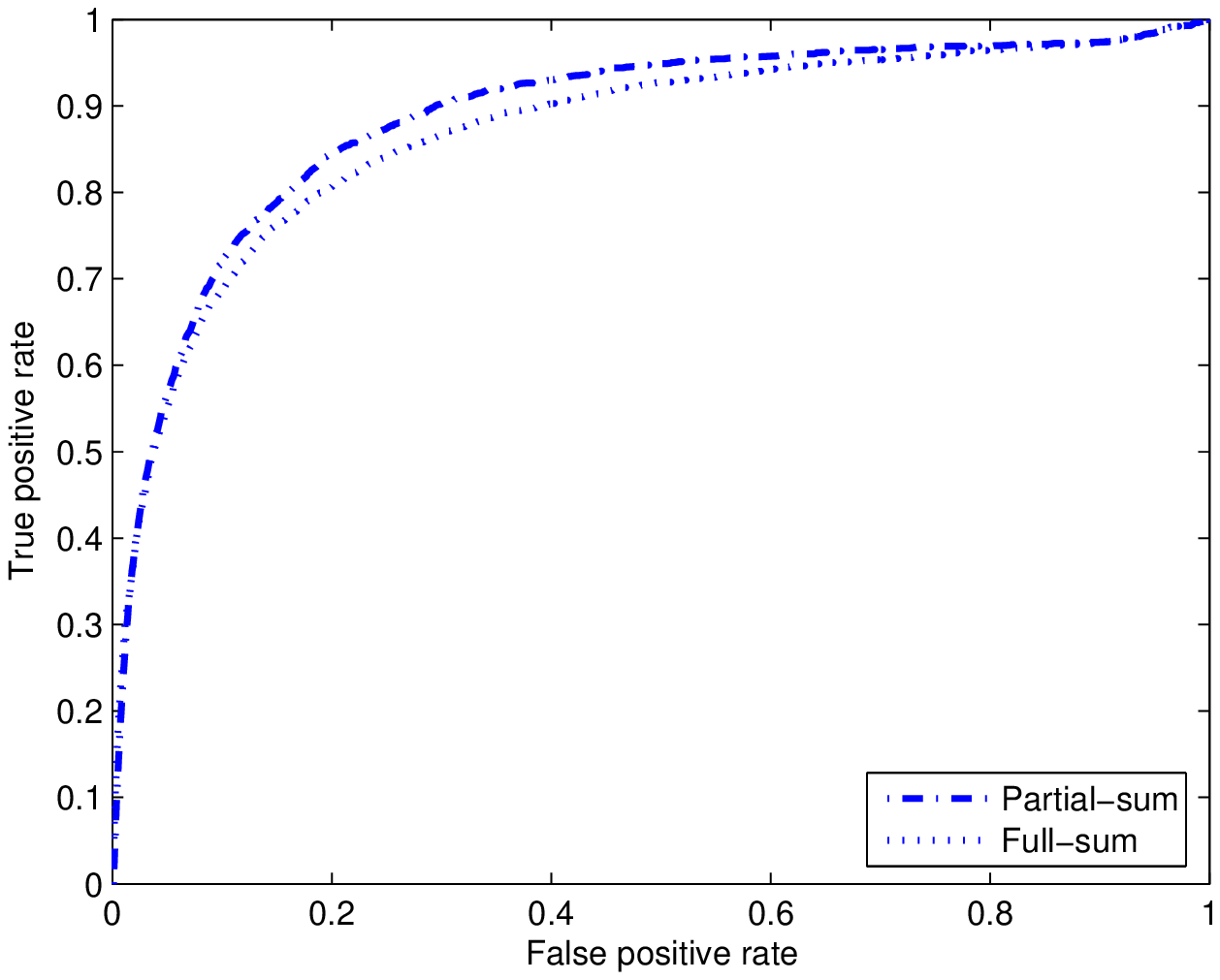}{5cm}{(b) $\alpha=0.5$ }{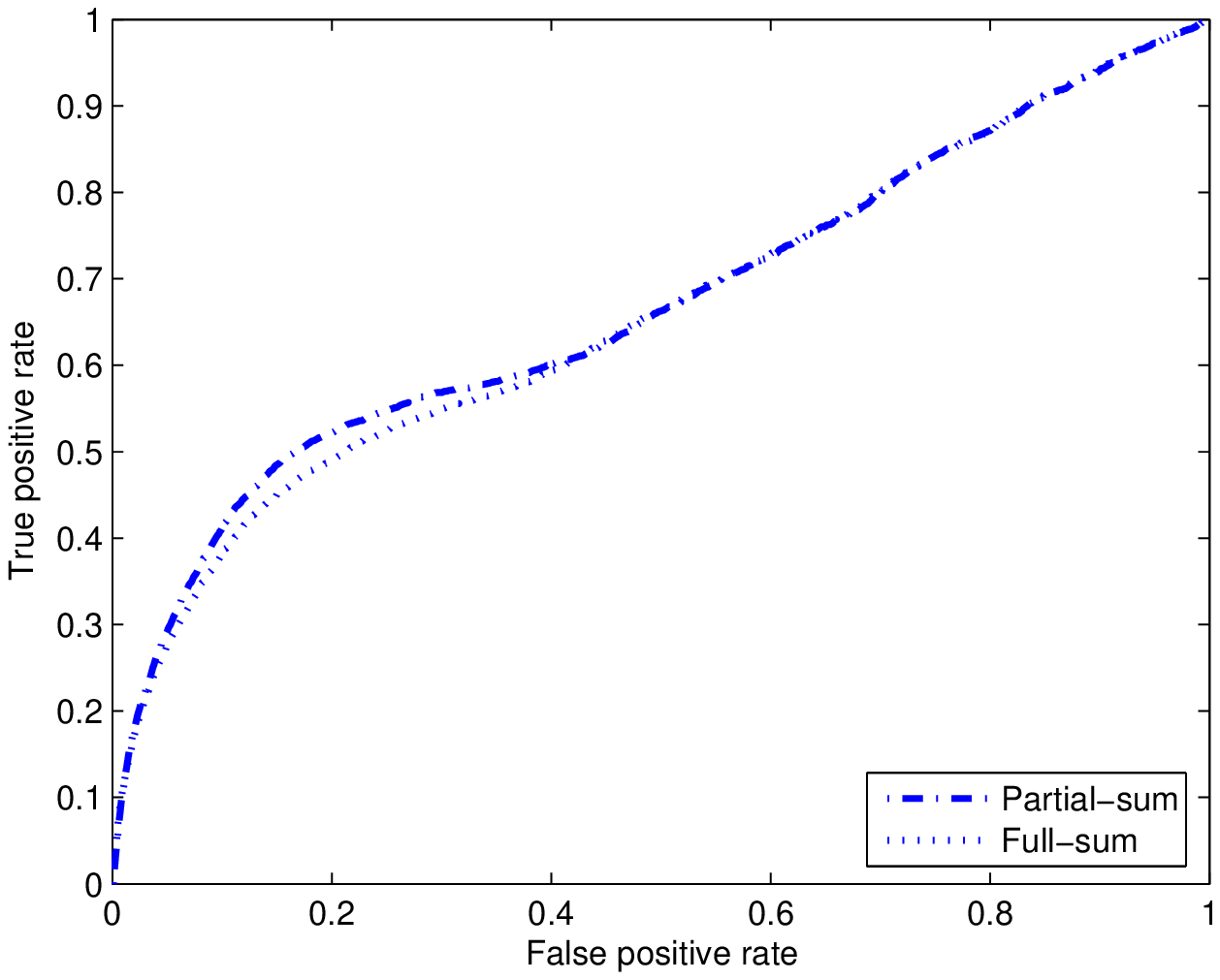}{5cm}{(c) $\alpha=0.2$ }
\caption{ROC curves for the school friendship network. }
\label{fig:school}
\end{figure} 

This dataset is a school friendship
network from the National Longitudinal Study of
Adolescent Health (see \cite{Hunter2008} for detailed information). This network contains 1011 high school students and 5459 directed links connecting students to their friends, as reported by the students themselves.  The average degree of this network is also around $5$.   Here we test our two link prediction criteria, with the same settings for $E_{ij}$ as in the protein example.  Since the latent variable model of \cite{Hoff2007} is not applicable to directed networks, we omit it here.     Due to lack of node covariates, we construct a network-based similarity $W$ by using the Jaccard index defined in \eqref{topo2}. We again apply block coordinate descent to minimize the criteria with $\lambda$ chosen by cross-validation, and report the average ROC curves over 10 realizations of $E_{ij}$'s. As shown in Figure \ref{fig:school}, both criteria perform fairly well for $\alpha=0.8$ and $\alpha=0.5$, but fail for $\alpha=0.2$, as the sampling rate is too small for $W$ to capture the overall network topology.   This does not happen in the protein-protein interactions network, since $W$ is constructed from covariates on proteins and is unaffected by sub-sampling.

\section{Summary and future work} \label{sec:summary}
In this article, we have proposed a new framework for link prediction that allows uncertainty in observed links and non-links of a given network. Our method can provide relative rankings of potential links for pairs with and  without observed links. The proposed link prediction criteria are fully non-parametric and essentially model-free, relying only on the assumption that similar node pairs have similar link probabilities, which is valid for a wide range of network models.   One direction we would like to explore in the future is to combine more specific parametric network models with our non-parametric approach, with the goal of achieving both robustness and efficiency.   We are also investigating consistency properties of our method, which is challenging because it requires developing a novel theoretical framework for evaluating consistency of rankings.   We are also developing extensions that would allow the probabilities of errors, $\alpha$ and $\beta$, to depend on the underlying probabilities of links.  This would allow, for example, making highly probable links more likely to be observed correctly.    Ultimately, we would also like to incorporate the general framework of link uncertainty into other network problems, for example, community detection.

\end{document}